\documentclass{article}

\usepackage{graphicx,color,algorithmic,algorithm} 

\usepackage{mlapa,amssymb,amsmath}

\newcounter{hyp}
\newenvironment{hyp}{\refstepcounter{hyp}\begin{itemize}
  \item[ \hspace*{.4cm} ({\bf{A}\arabic{hyp}})]}{\end{itemize}}

\newcommand{\hypref}[1]{({\bf{A}\ref{hyp:#1}})}
\newcommand{\hypreff}[2]{({\bf{A}\ref{hyp:#1}-\ref{hyp:#2}})}

\def \w{  {\mathbf{w}} }
\def \s{  {\mathbf{s}} }
\def \hw{  \hat{w} }
\def \J { {\mathbf{J} } }
  
\def \hq{  {q}}
\def \hQ{ {Q}}
\def \Q{ \mathbf{Q}}
\def \X{ \overline{X}}
\def \Y{ \overline{Y}}
\def \e{ \overline{\varepsilon}}
\def \lmin{ \lambda_{\min} }

\newcommand{\mysec}[1]{Section~\ref{sec:#1}}

\newcommand{\eq}[1]{Eq.~(\ref{eq:#1})}
\newcommand{\myfig}[1]{Figure~\ref{fig:#1}}

\newcommand{\BEAS}{\begin{eqnarray*}}
\newcommand{\EEAS}{\end{eqnarray*}}
\newcommand{\BEA}{\begin{eqnarray}}
\newcommand{\EEA}{\end{eqnarray}}
\newcommand{\BEQ}{\begin{equation}}
\newcommand{\EEQ}{\end{equation}}
\newcommand{\BIT}{\begin{itemize}}
\newcommand{\EIT}{\end{itemize}}
\newcommand{\BNUM}{\begin{enumerate}}
\newcommand{\ENUM}{\end{enumerate}}
\newcommand{\BA}{\begin{array}}
\newcommand{\EA}{\end{array}}

\newcommand{\var}{\mathop{\rm var}}

\newcommand{\sign}{\mathop{ \rm sign}}

\newcommand{\rb}{\mathbb{R}}

\def \E{{\mathbb E}}
\def \P{{\mathbb P}}

\newcommand{\BlackBox}{\rule{1.5ex}{1.5ex}}  

\newtheorem{proposition}{Proposition}
\newtheorem{lemma}{Lemma}

\title{Bolasso: Model Consistent Lasso Estimation through the Bootstrap}

\author{Francis R. Bach \\
francis.bach@mines.org \\
 INRIA - WILLOW Project-Team \\
Laboratoire d'Informatique de l'Ecole Normale Sup\'erieure \\
(CNRS/ENS/INRIA UMR 8548) \\
45 rue d'Ulm, 75230 Paris, France }

\begin{document} 
 
 \maketitle

\begin{abstract}
We consider the least-square linear regression problem with regularization by the $\ell_1$-norm, a problem usually referred to as the Lasso.
In this paper, we present a detailed asymptotic analysis of model consistency of the Lasso. For various decays of the regularization parameter, we compute asymptotic equivalents of the probability of correct model selection (i.e., variable selection). For a specific rate decay, we show that the Lasso selects all the variables that should enter the model with probability tending to one exponentially fast, while it selects all other variables with strictly positive probability. We show that this property implies that if we run the Lasso for several bootstrapped replications of a given sample, then intersecting the supports of the Lasso bootstrap estimates leads to consistent model selection. This novel variable selection algorithm, referred to as the Bolasso, is compared favorably to other linear regression methods on synthetic data and datasets from the UCI machine learning repository.

 \end{abstract} 
 
 \section{Introduction}
 Regularization by the $\ell_1$-norm has attracted a lot of interest in recent years in machine learning, statistics and signal processing. In the context of least-square linear regression, the problem is usually referred to as the \emph{Lasso}~\cite{lasso}.
Much of the early effort
 has been dedicated to algorithms to solve the optimization problem efficiently. In particular, the \emph{Lars}
 algorithm of~\singleemcite{lars} allows to find the entire regularization path (i.e., the set of solutions for all values
 of the regularization parameters) at the cost of a single matrix inversion.

 Moreover, a well-known justification of the 
regularization by the $\ell_1$-norm is that it leads to \emph{sparse} solutions, i.e., loading vectors with many zeros, and thus performs model selection. Recent works
\cite{Zhaoyu,yuanlin,zou,martin} have looked precisely
at the model consistency of the Lasso, i.e., if we know that the data were generated from a sparse loading vector, does
the Lasso actually recover it when the number of observed data points grows? In the case of a fixed number of covariates, the Lasso does recover
the sparsity pattern if and only if a certain simple condition on the generating covariance matrices is verified~\cite{yuanlin}. 
In particular, in low correlation settings,
the Lasso is indeed consistent. However, in presence of strong correlations, the Lasso cannot be consistent, shedding light on
potential problems of such procedures for variable selection. Adaptive versions where data-dependent weights
are added to the $\ell_1$-norm  then allow
to keep the consistency in all situations~\cite{zou}.

In this paper, we first derive a detailed asymptotic analysis of sparsity pattern selection of the Lasso estimation procedure, that extends previous analysis~\cite{Zhaoyu,yuanlin,zou}, by focusing on a specific decay of the regularization parameter. We show that when the decay is proportional to $n^{-1/2}$, where $n$ is the number of observations, then the Lasso will select all the variables that should enter the model (the \emph{relevant} variables) with probability tending to one exponentially fast with~$n$, while it selects all other variables (the \emph{irrelevant} variables) with strictly positive probability. If several datasets generated from the same distribution were available, then the latter property  would suggest to consider the intersection of the supports of the Lasso estimates for each dataset: all relevant variables would always be selected for all datasets, while irrelevant variables would enter the models randomly, and intersecting the supports from sufficiently many different datasets would simply eliminate them. However, in practice, only one dataset is given; but resampling methods such as the \emph{bootstrap} are exactly dedicated to mimic the availability of several datasets by resampling from the same unique dataset~\cite{efron}. In this paper, we show that when using the bootstrap and intersecting the supports, we actually get a consistent model estimate, without the consistency condition required by the regular Lasso. We refer to this new procedure as the \emph{Bolasso} (\textbf{bo}otstrap-enhanced \textbf{l}east \textbf{a}b\textbf{s}olute \textbf{s}hrinkage \textbf{o}perator).  Finally, our Bolasso framework could be seen as a voting scheme applied to the supports of the bootstrap Lasso estimates;
however, our procedure may rather be considered as a consensus combination scheme,
as we keep the (largest) subset of variables on which \emph{all} regressors agree in terms of variable selection, which is in our case provably consistent and also allows to get rid of a potential additional hyperparameter.

The paper is organized as follows: in \mysec{analysis}, we present the asymptotic analysis of model selection for the Lasso; in \mysec{bootstrap}, we describe the Bolasso algorithm as well as its proof of model consistency, while in \mysec{simulations}, we illustrate our results on synthetic data, where the true sparse generating model is known, and data from the UCI machine learning repository. Sketches of proofs can be found in Appendix~A.

 \paragraph{Notations}
 For a vector $v \in \rb^p$, we let denote $\| v\|_2 = (v^\top v)^{1/2}$ the $\ell_2$-norm, $\| v\|_\infty = \max_{i \in \{1,\dots,p\} }|v_i|$ the $\ell_\infty$-norm and $\| v\|_1 = \sum_{i=1}^p |v_i|$ the $\ell_1$-norm. For   $a \in \rb$, $\sign(a)$ denotes the sign of $a$, defined as $\sign(a)=1$ if $a>0$, $-1$ if $a<0$, and $0$ if $a=0$. For a vector $v \in \rb^p$, $\sign(v) \in \rb^p$ denotes the the vector of signs of elements of $v$.
 
 Moreover, given a vector $v \in \rb^p$ and a subset $I$ of $\{1,\dots,p\}$, $v_I$ denotes the vector in $\rb^{{\rm Card}(I)}$ of elements of $v$ indexed by $I$. Similarly, for a matrix $A \in \rb^{p \times p}$, $A_{I,J}$ denotes the submatrix of  $A$ composed of elements of $A$ whose rows are in $I$ and columns are in $J$.
 
 \section{Asymptotic Analysis of Model Selection for the Lasso}
 \label{sec:analysis}
 
 In this section, we describe existing and new asymptotic results regarding the model selection capabilities of the Lasso.
 
 \subsection{Assumptions}
 
 We consider the problem of predicting a response $Y \in \rb$ from covariates $X= (X_1,\dots,X_p)^\top \in \rb^{p}$.
The only assumptions that we make on the joint distribution $P_{XY}$ of $(X,Y)$ are the following:

\begin{hyp}
\label{hyp:var}
The cumulant generating functions
$\E \exp( s\|X\|_2^2 ) $ and $\E\exp( s Y^2)$ are finite for some $s>0$.
\end{hyp}

\begin{hyp}
\label{hyp:inv}
 The joint matrix of second order moments   $\Q = \E XX^\top  \in \rb^{ p \times p} $ is invertible.
\end{hyp}

\begin{hyp}
\label{hyp:model}
 $\E ( Y | X )  = X^\top \w $ and $\var(Y|X) = \sigma^2 \mbox{ a.s.} $ for some $\w \in \rb^p$ and $\sigma \in \rb_+^\ast$.  
\end{hyp}

We let denote $\J = \{ j, \w_j \neq 0\}$
the sparsity pattern of $\w$, $\s=\sign(\w)$ the sign pattern of $\w$, and $\varepsilon = Y - X^\top \w$ the additive noise.\footnote{
 Throughout this paper, we use boldface fonts for population quantities.
} Note that our assumption regarding cumulant generating functions is satisfied when $X$ and $\varepsilon$ have compact support, and also, when the densities of $X$ and $\varepsilon$ have light tails.

We consider \emph{independent and identically distributed}
 (i.i.d.) data $(x_i,y_i) \in \rb^p \times \rb$, $i=1,\dots,n$, sampled from $P_{XY}$;
the data are given in the form of matrices $\Y \in \rb^n$ and $\X \in \rb^{n \times p}$. 

Note that the i.i.d.~assumption, together with \hypreff{var}{model}, are the simplest assumptions for studying the asymptotic behavior of the Lasso; and it is of course of interest to allow more general assumptions, in particular growing number of variables $p$, more general random variables, etc. (see, e.g., \singleemcite{yuinfinite}), which are outside the scope of this paper.

\subsection{Lasso Estimation}

We consider the square loss function $\frac{1}{2n}
\sum_{i=1}^n ( y_i - w^\top x_i)^2 = \frac{1}{2n} \| \Y  - \X w \|_2^2$ and the regularization by the $\ell_1$-norm defined as $\|w\|_1 = \sum_{i=1}^p  | w_i|$. That is, we look at the following Lasso optimization problem~\cite{lasso}:
\BEQ
\label{eq:lasso} \min_{w \in \rb^p}  \textstyle\frac{1}{2n} \|\Y  - \X w \|_2^2 + \mu_n  \| w\|_1,
\EEQ
where $\mu_n \geqslant 0$ is the regularization parameter.
We   denote $\hw$ any global minimum of \eq{lasso}---it may not be unique in general, but will with probability tending to one exponentially fast under assumption \hypref{var}. 

\subsection{Model Consistency - General Results}

In this section, we detail the asymptotic behavior of the Lasso estimate $\hat{w}$, both in terms of the difference in norm with the population value $\w$ (i.e., regular
consistency) and of the \emph{sign pattern} $\sign(\hat{w})$, for all asymptotic behaviors of the regularization parameter $\mu_n$. Note that information about the sign pattern includes information about the \emph{support}, i.e., the indices $i$ for which $\hat{w}_i$ is different from zero; moreover, when $\hat{w}$ is consistent, consistency of the sign pattern is in fact equivalent to the consistency of the support.  

We now consider five mutually exclusive possible situations which explain various portions of the regularization path (we assume 
\hypreff{var}{model}); many of these results appear elsewhere~\cite{yuanlin,Zhaoyu,fu,zou,grouplasso} but some of the finer results presented below are new (see \mysec{detailed}). 
\BNUM
\item
If $\mu_n$ tends to infinity, then $\hat{w}=0$ with probability tending to one.
\item
If $\mu_n$ tends to a finite strictly positive constant $\mu_0$, then $\hat{w}$ converges in probability to the unique
global minimum of $\frac{1}{2} (w-\w)^\top \Q (w-\w) + \mu_0 \| w\|_1$. Thus, the estimate $\hat{w}$ never converges in probability to $\w$, while the sign pattern tends to the one of the previous global minimum, which may or may not be the same as the one of $\w$.\footnote{Here and in the third regime, we do not take into account the pathological cases where the sign pattern of the limit in unstable, i.e., the limit is exactly at a hinge point of the regularization path.}
\item If $\mu_n$ tends to zero slower than $n^{-1/2}$, then $\hat{w}$ converges in probability to $\w$ (regular consistency) and the sign pattern converges to the sign pattern of the global minimum of
$\frac{1}{2} v^\top \Q v + v_\J^\top \sign(\w_\J) + \| v_{\J^c}\|_1$. This sign pattern is equal to the population sign vector $\s=\sign(\w)$ if and only if the following consistency condition is satisfied:
\BEQ
\label{eq:cond}
 \| \Q_{\J^c \J} \Q_{\J \J}^{-1} \sign(\w_\J) \|_\infty \leqslant 1. 
 \EEQ
Thus, if \eq{cond} is satisfied, the probability of correct sign estimation is tending to one, and to zero otherwise~\cite{yuanlin}.
\item If $\mu_n = \mu_0 n^{-1/2}$ for $\mu_0 \in (0,\infty)$, then the sign pattern of $\hat{w}$ agrees on $\J$ with the one of $\w$  with probability tending to one, while for all sign patterns consistent on $\J$ with the one of $\w$, the probability of obtaining this pattern is tending to a limit in $(0,1)$ (in particular strictly positive); that is, all patterns consistent on $\J$ are possible with positive probability. See \mysec{detailed} for more details.
\item If $\mu_n$ tends to zero faster than $n^{-1/2}$, then $\hat{w}$ is consistent (i.e., converges in probability to $\w$) but the support of $\hat{w}$ is equal to $\{1,\dots,p\}$ with probability tending to one (the signs of variables in $\J^c$ may be negative or positive). That is, the $\ell_1$-norm has no sparsifying effect.
\ENUM

Among the five previous regimes, the only ones with consistent estimates (in norm) and a sparsity-inducing effect are $\mu_n$ tending to zero and $\mu_n n^{1/2}$ tending to a limit $\mu_0 \in (0,\infty]$ (i.e., potentially infinite). When $\mu_0 = +\infty$, then
we can only hope for model consistent estimates if the consistency condition in \eq{cond} is satisfied. This somewhat disappointing result for the Lasso has led to various improvements on the Lasso to ensure model consistency even when \eq{cond} is not satisfied~\cite{yuanlin,zou}. Those are based on adaptive weights based on the non regularized least-square estimate. We propose in \mysec{bootstrap} an alternative
 way which is based on resampling.
 
In this paper, we now consider the specific case where $\mu_n = \mu_0 n^{-1/2}$ for $\mu_0 \in (0,\infty)$, where we derive new asymptotic results. Indeed, in this situation, we get the correct signs of the relevant variables (those in $\J$) with probability tending to one, but we also get all possible sign patterns consistent with this, i.e., all other variables (those not in $\J$) may be non zero with asymptotically strictly positive probability. However, if we were to repeat the Lasso estimation for many datasets obtained from the same distribution, we would obtain for each $\mu_0$, a set of active variables, all of which include $\J$ with probability tending to one, but potentially containing all other subsets. By intersecting those, we would get exactly~$\J$. 

However, this requires multiple copies of the samples, which are not usually available. Instead, we consider bootstrapped samples which exactly mimic the behavior of having multiple copies. See \mysec{bootstrap} for more details.

\subsection{Model Consistency with Exact Root-$n$ Regularization Decay }
\label{sec:detailed}
In this section we present detailed new results regarding the pattern consistency for
$\mu_n$ tending to zero exactly at rate $n^{-1/2}$ (see proofs in Appendix A):
\begin{proposition}
\label{prop:prop1}
Assume  \hypreff{var}{model} and  $\mu_n=\mu_0 n^{-1/2}$, $\mu_0>0$. Then for any sign pattern $s\in \{-1,0,1\}^p$ such that $s_\J = \sign(\w_\J)$,   $ \P( \sign(\hat{w}) = s )$ tends to a limit $\rho(s,\mu_0) \in (0,1)$, and we have:
$$\P( \sign(\hat{w}) = s ) - \rho(s,\mu_0) = O(n^{-1/2} \log n ).$$
\end{proposition}
\begin{proposition}
\label{prop:prop2}
Assume  \hypreff{var}{model} and  $\mu_n=\mu_0 n^{-1/2}$, $\mu_0>0$. Then,
 for any pattern $s\in \{-1,0,1\}^p$ such that  $s_\J \neq \sign(\w_\J)$,   there exist a constant $A(\mu_0)>0$ such that 
$$
 \log \P( \sign(\hat{w}) = s )  \leqslant -nA(\mu_0)  + O(n^{-1/2}) . $$
\end{proposition}
The last two propositions state that we get all relevant variables with probability tending to one \emph{exponentially fast}, while we get exactly get all other patterns with probability tending to a limit
\emph{strictly} between zero and one. Note that the results that we give in this paper are valid for \emph{finite} $n$, i.e., we could derive actual bounds on probability of sign pattern selections with known constants that explictly depend on~$\w$,~$\Q$ and~$P_{XY}$.

\section{Bolasso: Bootstrapped Lasso}
\label{sec:bootstrap}
Given the $n$ i.i.d.~observations $(x_i,y_i) \in \rb^d \times \rb$, $i=1,\dots,n$, given by matrices
$\X \in \rb^{ n \times p}$ and $\Y \in \rb^n$, we consider $m$ \emph{bootstrap} replications of the $n$ data points~\cite{efron}; that is, for $k=1,\dots,m$, we consider a \emph{ghost sample}
$(x^{k}_i,y^k_i)\in \rb^d \times \rb$, $i=1,\dots,n$, given by matrices
$\X^k \in \rb^{ n \times p}$ and $\Y^k \in \rb^n$. The $n$ pairs $(x^{k}_i,y^k_i)$,
$i=1,\dots,n$, are sampled uniformly at random \emph{with replacement} from the $n$ original pairs in $(\X,\Y)$. The sampling of the $nm$ pairs of observations is independent.
In other words, we defined the distribution of the  ghost sample $(\X^\ast,\Y^\ast)$
by sampling $n$ points with replacement from $(\X,\Y)$, and, given $(\X,\Y)$, the $m$ ghost samples are independently sampled i.i.d.~from the distribution of   $(\X^\ast,\Y^\ast)$.

The asymptotic analysis from \mysec{analysis} suggests to estimate the supports $J_k = \{
j,\ \hat{w}^k_j \neq 0\}$ of the Lasso estimates $\hat{w}^k$ for the bootstrap samples, $k=1,\dots,m$, and to intersect them to define the Bolasso model estimate of the support:
$J = \bigcap_{k=1}^m J_k$. Once $J$ is selected, we estimate $w$ by the unregularized least-square fit restricted to variables in $J$. The detailed algorithm is   
given in Algorithm~\ref{alg:bolasso}. The algorithm has only one extra parameter (the number of bootstrap samples $m$). Following Proposition~\ref{prop:bolasso}, $\log(m)$ should be chosen growing with $n$ asymptotically slower than $n$. In simulations, we always use $m=128$ (except in \myfig{figure1}, where we exactly study the influence of $m$).

\begin{algorithm}
   \caption{Bolasso}
   \label{alg:bolasso}
\begin{algorithmic}
   \STATE {\bfseries Input:} data $(\X,\Y) \in \rb^{n\times (p+1)}$ \\
   \hspace*{1.14cm}
   number of bootstrap replicates $m$ \\
   \hspace*{1.14cm}
   regularization parameter $\mu$
    \FOR{$k=1$ {\bfseries to} $m$}
   \STATE  Generate bootstrap samples $(\X^k,\Y^k) \in \rb^{n\times (p+1)}$ 
   \STATE Compute Lasso estimate $\hat{w}^k$ from $(\X^k,\Y^k)$
   \STATE Compute support $J_k = \{ j,\ \hat{w}^k_j \neq 0\}$
     \ENDFOR
     \STATE Compute  $J = \bigcap_{k=1}^m J_k$
     \STATE Compute $\hat{w}_J$ from $(\X_{J},\Y) $
 \end{algorithmic}
\end{algorithm}

Note that in practice, the Bolasso estimate can be computed simultaneously for a large number of regularization parameters because of the efficiency of the Lars algorithm (which we use in simulations), that allows to find the entire regularization path for the Lasso at the (empirical) cost of a single matrix inversion~\cite{lars}. Thus computational complexity of the Bolasso is $O(m(p^3+p^2n))$.

The following proposition (proved in Appendix~A) shows that the previous algorithm leads to consistent model selection. 
\begin{proposition}
\label{prop:bolasso}
Assume  \hypreff{var}{model} and  $\mu_n=\mu_0 n^{-1/2}$, $\mu_0>0$. Then the probability that the Bolasso does not exactly select the correct model, i.e., for all $m>0$,
$\P(J \neq \J)$ has the following upper bound:
$$ \textstyle \P(J \neq \J) \leqslant m A_1 e^{- A_2 n} + A_3 \frac{\log n}{n^{1/2}} + A_4\frac{\log m }{m},$$
where $A_1, A_2, A_3, A_4$ are strictly positive constants.
\end{proposition}

 Therefore, if $\log(m)$ tends to infinity slower than $n$ when $n$ tends to infinity, the Bolasso asymptotically  selects  with overwhelming probability the correct active variable, and by regular consistency of the restricted least-square estimate, the correct sign pattern as well. Note that
the previous bound is true whether the condition in \eq{cond} is satisfied or not, but could be improved on if we suppose that \eq{cond} is satisfied. See \mysec{synthetic} for a detailed comparison with the Lasso on synthetic examples.

\section{Simulations}
\label{sec:simulations}
In this section, we illustrate the consistency results obtained in this paper with a few simple simulations on
synthetic examples similar to the ones used by~\singleemcite{grouplasso} and some medium scale datasets from the UCI machine learning repository~\cite{UCI}.

\subsection{Synthetic examples}
\label{sec:synthetic}
For a given dimension $p$, we  sampled $X \in \rb^p$ from a normal distribution with zero mean and covariance matrix   generated as follows: (a) sample a $p\times p$ matrix $G$ with independent standard normal distributions, (b) form $\Q = GG^\top$, (c) scale $\Q$ to unit diagonal.
We then selected the first ${\rm Card} ( \J) = r$ variables and sampled non zero loading vectors as follows: (a) sample each loading from  independent standard normal distributions,  (b) rescale those to unit magnitude, (c) rescale those by a scaling which is uniform at random between $\frac{1}{3}$ and $1$ (to ensure $\min_{j \in \J} | \w_j| \geqslant 1/3$). Finally, we chose a constant noise level $\sigma$ equal to
$0.1$ times $(\E (\w^\top X)^2)^{1/2}$, and the additive noise $\varepsilon$ is normally distributed with zero mean and variance $\sigma^2$.
Note that the joint distribution on $(X,Y)$ thus defined satisfies with probability one (with respect to the sampling of the covariance matrix) assumptions \hypreff{var}{model}.

\begin{figure}
\begin{center}
\hspace*{-.35cm}
\includegraphics[width=4.15cm, height=3.25cm]{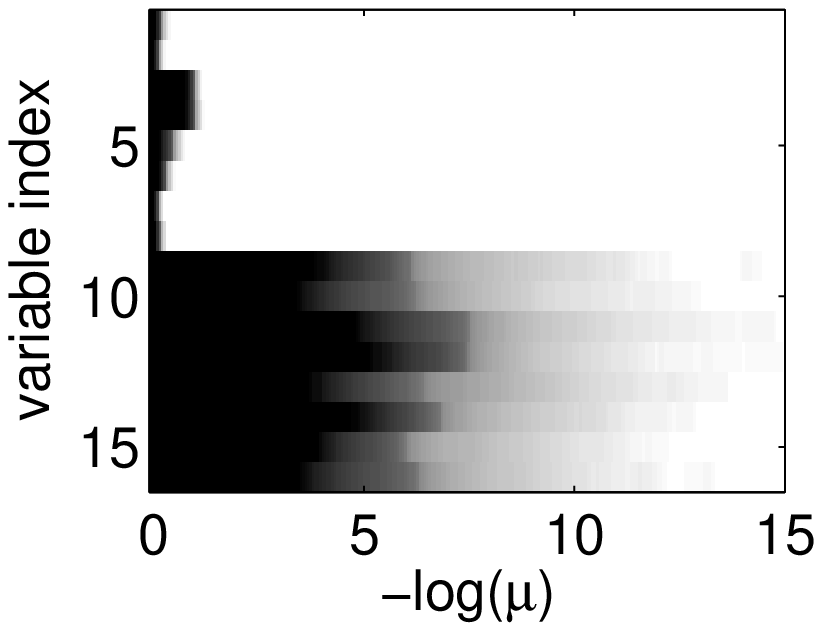}
\hspace*{-.1cm}
\includegraphics[width=4.15cm, height=3.25cm]{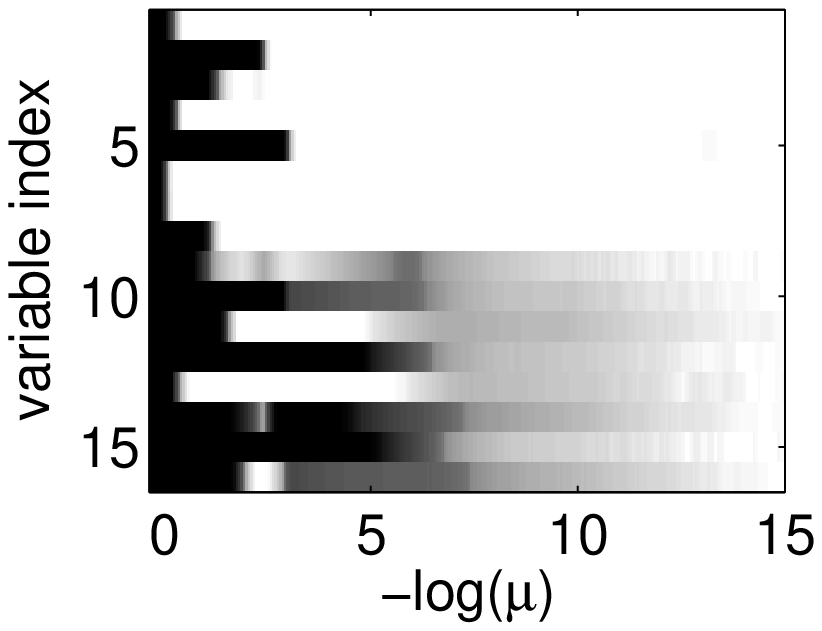}
\hspace*{-.1cm}
\end{center}
 
\caption{\textbf{Lasso}: log-odd ratios of the probabilities of selection of each variable (white = large probabilities, black = small probabilities) vs. regularization parameter. Consistency condition in \eq{cond} satisfied (left) and not satisfied (right).}
\label{fig:figure2}
\end{figure}

\begin{figure}
\begin{center}
\hspace*{-.35cm}
\includegraphics[width=4.15cm, height=3.25cm]{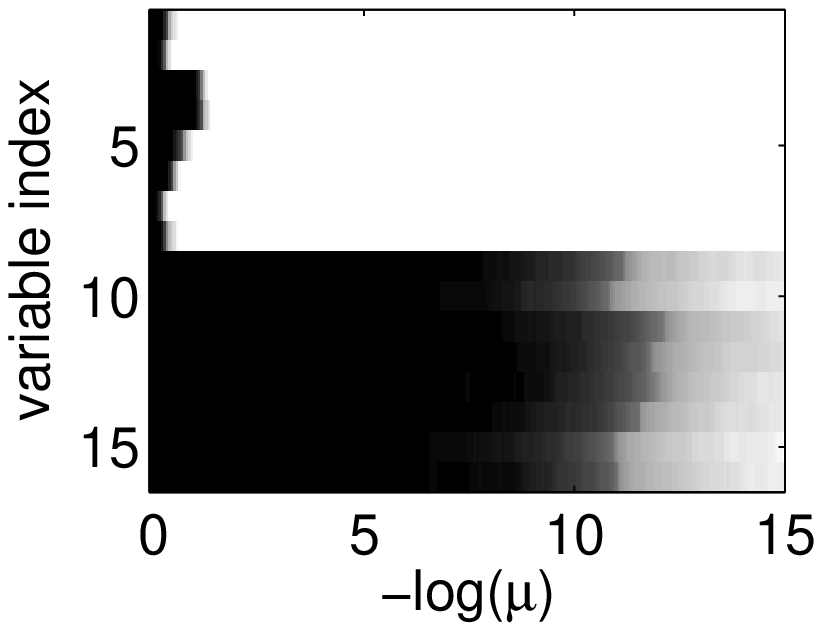}
\hspace*{-.1cm}
\includegraphics[width=4.15cm, height=3.25cm]{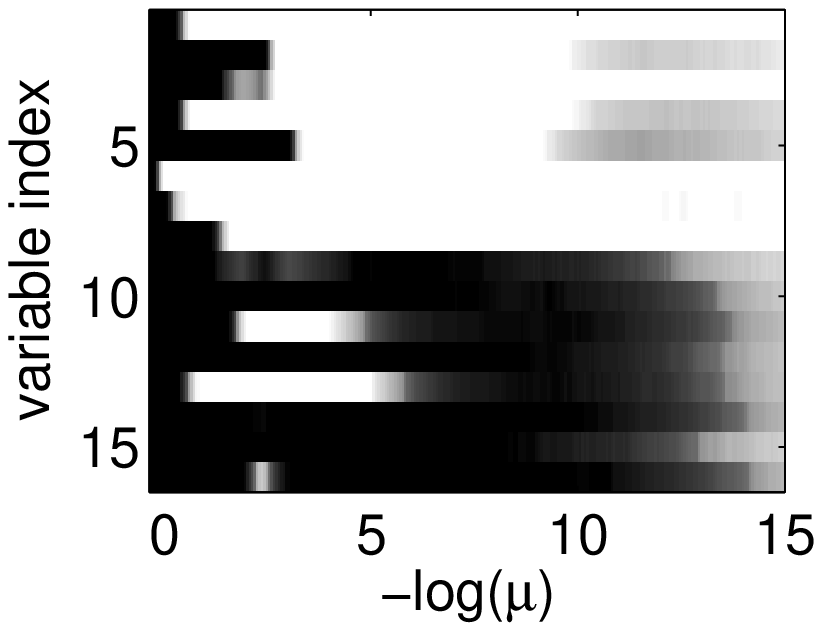}
\hspace*{-.1cm}
\end{center}
 
\caption{\textbf{Bolasso}: log-odd ratios of the probabilities of selection of each variable (white = large probabilities, black = small probabilities) vs. regularization parameter. Consistency condition in \eq{cond} satisfied (left) and not satisfied (right).}
\label{fig:figure2bolasso}
\end{figure}

In \myfig{figure2}, we sampled two   distributions $P_{XY}$ with $p=16$ and
$r=8$ relevant variables, one for which the consistency condition in \eq{cond} is satisfied (left), one for which it was not satisfied (right). For a fixed number of sample $n=1000$, we generated 256 replications and computed the empirical frequencies of selecting any given variable for the Lasso as the regularization parameter $\mu$ varies. Those plots show the various asymptotic regimes of the Lasso detailed in \mysec{analysis}. In particular, on the right plot, although no $\mu$ leads to perfect selection (i.e., exactly variables with indices less than $r=8$ are selected), there is a range where all relevant variables are always selected, while all others are  selected with probability within $(0,1)$.  

In \myfig{figure2bolasso}, we plot the results under the same conditions for the Bolasso (with a fixed number of bootstrap replications $m=128$). We can see that in the Lasso-consistent case (left), the Bolasso widens the consistency region, while in the Lasso-inconsistent case (right), the Bolasso ``creates'' a consistency region.

\begin{figure}
\begin{center}
\hspace*{-.35cm}
\includegraphics[width=4.15cm, height=3.25cm]{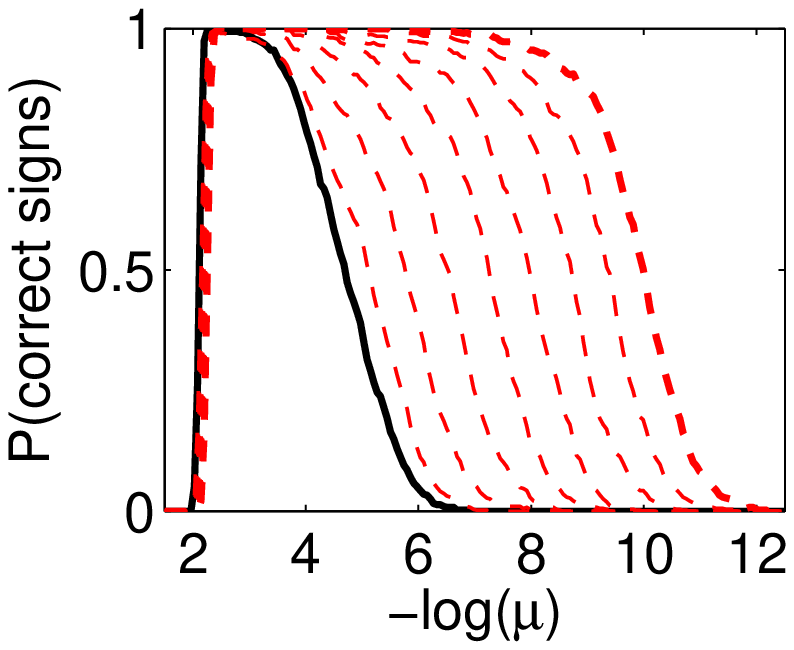}
\hspace*{-.1cm}
\includegraphics[width=4.15cm, height=3.25cm]{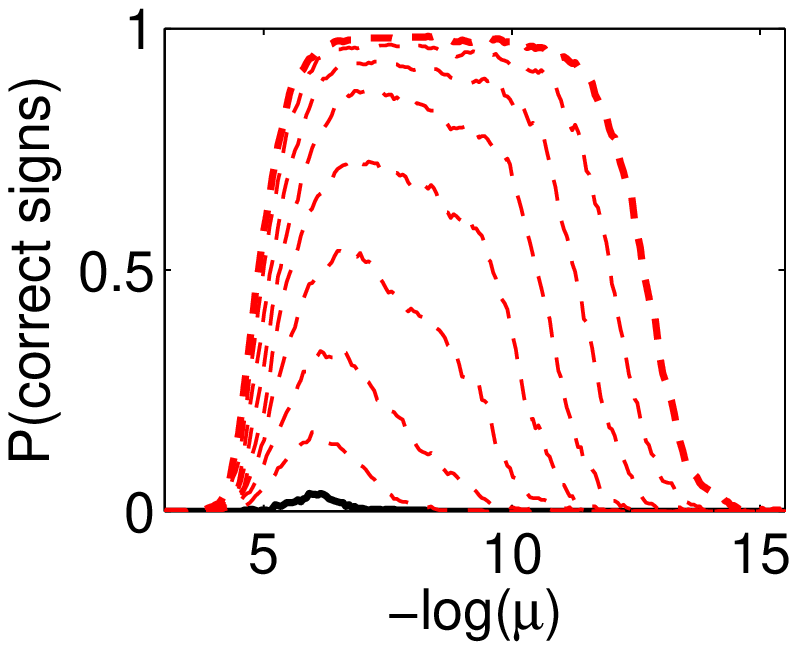}
\hspace*{-.1cm}
\end{center}
 
\caption{Bolasso (red, dashed) and Lasso (black, plain): probability of correct sign estimation vs. regularization parameter. Consistency condition in \eq{cond} satisfied (left) and not satisfied (right). The number of bootstrap replications $m$ is in $\{2,4,8,16,32,64,128,256\}$. }
\label{fig:figure1}
\end{figure}

 In \myfig{figure1}, we  selected the same two distributions and compared the probability of exactly selecting the correct support pattern, for the Lasso, and for the Bolasso with varying numbers of bootstrap replications (those probabilities are computed by averaging over 256 experiments with the same distribution). In \myfig{figure1}, we can see that in the Lasso-inconsistent case (right), the Bolasso indeed allows to  fix the unability of the Lasso to find the correct pattern. Moreover, increasing $m$ looks always beneficial; note that although it seems to contradict the asymptotic analysis in \mysec{bootstrap} (which imposes an upper bound for consistency), this is due to the fact that not selecting (at least) the relevant variables has very low probability and is not observed with only 256 replications.

\begin{figure}
\begin{center}
\hspace*{-.35cm}
\includegraphics[width=4.15cm, height=3.25cm]{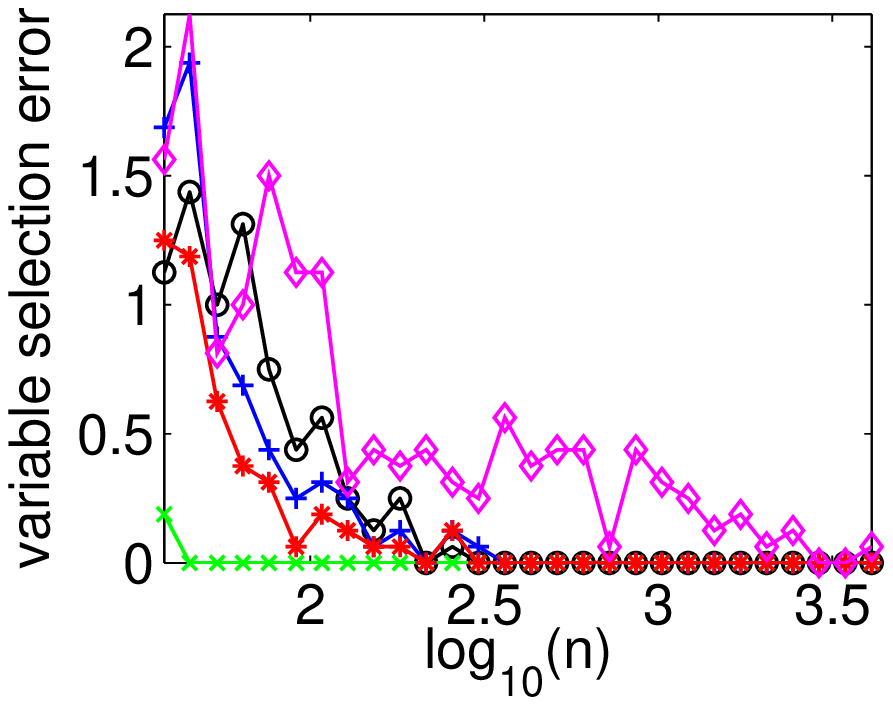}
\hspace*{-.1cm}
\includegraphics[width=4.15cm, height=3.25cm]{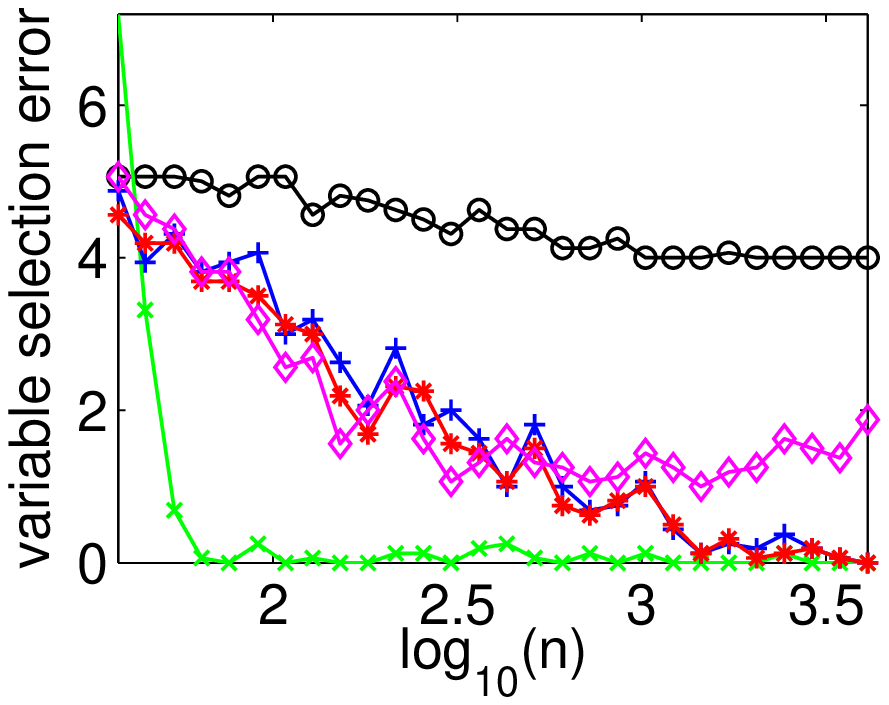}
\hspace*{-.1cm}
\end{center}
 
\caption{Comparison of several variable selection methods: Lasso (black circles),
Bolasso (green crosses), forward greedy (magenta diamonds), thresholded LS estimate (red stars), adaptive Lasso (blue pluses). Consistency condition in \eq{cond} satisfied (left) and not satisfied (right). The averaged (over 32 replications) variable selection error is computed as the square distance between sparsity pattern indicator vectors.}
\label{fig:figure4}
\end{figure}

Finally, in \myfig{figure4}, we compare various variable selection procedures for linear regression, to the Bolasso, with two distributions where $p=64$, $r=8$ and varying $n$.
 For all the methods we consider, there is a natural way to select exactly $r$ variables with no free parameters (for the Bolasso, we select the most stable pattern with $r$ elements, i.e., the pattern which corresponds to most values of $\mu$). We can see that the Bolasso outperforms all other variable selection methods, even in settings where the number of samples becomes of the order of the number of variables, which requires additional theoretical analysis, subject of ongoing research. Note in particular that we compare with bagging of least-square regression~\cite{bagging} followed by a thresholding of the loading vector, which is another simple way of using bootstrap samples: the Bolasso provides a more efficient way to use the extra information, not for usual stabilization purposes~\cite{stabilizing}, but directly for model selection. Note finally, that the bagging of Lasso estimates requires an additional parameter and is thus not tested.

\subsection{UCI datasets}
The previous simulations have shown that the Bolasso is succesful at performing model selection in synthetic examples. We now apply it to several linear regression problems and compare it to alternative methods for linear regression, namely, ridge regression, Lasso, bagging of Lasso estimates~\cite{bagging}, and a soft version of the Bolasso (referred to as Bolasso-S), where instead of intersecting the supports for each bootstrap replications, we select those which are present in at least $90\%$ of the bootstrap replications. In Table~\ref{table1}, we consider data randomly generated as in \mysec{synthetic} (with $p=32$, $r=8$, $n=64$), where the true model is known to be composed of a sparse loading vector, while in Table~\ref{table2}, we consider regression datasets from the UCI machine learning repository. For all of those, we perform 10 replications of 10-fold cross validation and for all methods (which all have one free regularization parameter), we select the best regularization parameter on the 100 folds and plot the mean square \emph{prediction} error and its standard deviation.

Note that when the generating model is actually sparse (Table~\ref{table1}), the Bolasso outperforms all other models, while in other cases (Table~\ref{table2}) the Bolasso is sometimes too strict in intersecting models, i.e., the softened version works better and is competitive with other methods. Studying the effects of this softened scheme (which is more similar to usual voting schemes), in particular in terms of the potential trade-off between good model selection and low prediction error, and under conditions where $p$ is large, is the subject of ongoing work.

   \section{Conclusion}
  We have presented a detailed analysis of variable selection properties of a boostrapped version of the Lasso. The model estimation procedure, referred
  to as the Bolasso, is provably consistent under general assumptions. 
  This work brings to light that poor variable selection results of the Lasso may be easily enhanced  thanks to a simple parameter-free resampling procedure. 
Our contribution also suggests that the use of bootstrap samples by L. Breiman
in Bagging/Arcing/Random Forests~\cite{arcing} may have been so far slightly overlooked and considered a minor feature, while using boostrap samples may actually be 
a key computational feature in such algorithms for good model selection performances, and eventually good prediction performances on real datasets.
  
  The current work could be extended in various ways: first, we have focused on a fixed total number of variables, and allowing the numbers
  of variables to grow is important in theory and in practice~\cite{yuinfinite}. Second, the same technique can be applied to similar settings than least-square regression with the $\ell_1$-norm, namely regularization by block $\ell_1$-norms~\cite{grouplasso} and other losses such as general convex classification losses. Finally, theoretical and practical connections could be made with other work on resampling methods and boosting~\cite{boosting}.

\begin{table}
\caption{Comparison of least-square estimation methods, data generated as described in \mysec{synthetic}, with $\kappa = \| \Q_{\J^c\J} \Q_{\J\J}^{-1} \s_\J\|_\infty$ (cf. \eq{cond}). Performance is measured through mean squared prediction error (multiplied by 100).}

\label{table1}

\vspace*{.05cm}

\begin{center}

\begin{tabular}{|l|l|l|l|l|}
\hline
 $\kappa$ &   0.93 &    1.20 &    1.42  &   1.28 \\
\hline
 \hspace*{-.25cm} Ridge     \hspace*{-.25cm}     &  $ \!  8.8 \pm 4.5 \! $   &  $ \!  4.9 \pm 2.5 \! $   &  $ \!  7.3 \pm 3.9 \! $   &  $ \!  8.1 \pm 8.6 \! $   \\ 
 \hspace*{-.25cm} Lasso     \hspace*{-.25cm}     &  $ \!  7.6 \pm 3.8 \! $   &  $ \!  4.4 \pm 2.3 \! $   &  $ \!  4.7 \pm 2.5 \! $   &  $ \!  5.1 \pm 6.5 \! $   \\ 
 \hspace*{-.25cm} Bolasso   \hspace*{-.25cm}     &  $ \!  5.4 \pm 3.0 \! $   &  $ \!  3.4 \pm 2.4 \! $   &  $ \!  3.4 \pm 1.7 \! $   &  $ \!  3.7 \pm 10.2 \! $   \\ 
 \hspace*{-.25cm} Bagging   \hspace*{-.25cm}     &  $ \!  7.8 \pm 4.7 \! $   &  $ \!  4.6 \pm 3.0 \! $   &  $ \!  5.4 \pm 4.1 \! $   &  $ \!  5.8 \pm 8.4 \! $   \\ 
 \hspace*{-.25cm} Bolasso-S \hspace*{-.25cm}     &  $ \!  5.7 \pm 3.8 \! $   &  $ \!  3.0 \pm 2.3 \! $   &  $ \!  3.1 \pm 2.8 \! $   &  $ \!  3.2 \pm 8.2 \! $   \\ 
 \hline
\end{tabular}
 \end{center}

\end{table}

\begin{table}
\caption{Comparison of least-square estimation methods, UCI regression datasets. Performance is measured through mean squared prediction error (multiplied by 100).}
\label{table2}

\vspace*{.05cm}
\begin{center}

\begin{tabular}{|l|l|l|l|l|}
\hline
& \hspace*{-.2cm}  Autompg \hspace*{-.3cm} &  Imports & Machine & Housing \\ 
\hline
 \hspace*{-.25cm} Ridge     \hspace*{-.25cm}     & $ \! 18.6 \! \pm \! 4.9   \! $  & $ \! 7.7 \! \pm \! 4.8   \! $  & $ \! 5.8 \! \pm \! 18.6   \! $  & $ \! 28.0 \! \pm \! 5.9   \! $  \\ 
 \hspace*{-.25cm} Lasso     \hspace*{-.25cm}     & $ \! 18.6 \! \pm \! 4.9   \! $  & $ \! 7.8 \! \pm \! 5.2   \! $  & $ \! 5.8 \! \pm \! 19.8   \! $  & $ \! 28.0 \! \pm \! 5.7   \! $  \\ 
 \hspace*{-.25cm} Bolasso   \hspace*{-.25cm}     & $ \! 18.1 \! \pm \! 4.7   \! $  & $ \! 20.7 \! \pm \! 9.8   \! $  & $ \! 4.6 \! \pm \! 21.4   \! $  & $ \! 26.9 \! \pm \! 2.5   \! $  \\ 
 \hspace*{-.25cm} Bagging   \hspace*{-.25cm}     & $ \! 18.6 \! \pm \! 5.0   \! $  & $ \! 8.0 \! \pm \! 5.2   \! $  & $ \! 6.0 \! \pm \! 18.9   \! $  & $ \! 28.1 \! \pm \! 6.6   \! $  \\ 
 \hspace*{-.25cm} Bolasso-S \hspace*{-.25cm}     & $ \! 17.9 \! \pm \! 5.0   \! $  & $ \! 8.2 \! \pm \! 4.9   \! $  & $ \! 4.6 \! \pm \! 19.9   \! $  & $ \! 26.8 \! \pm \! 6.4   \! $  \\ 
\hline
\end{tabular}

\end{center}

 \end{table}

 \appendix
  \section{Proof of Model Consistency Results}
In this appendix, we give sketches of proofs for the asymptotic
 results presented in \mysec{analysis} and \mysec{bootstrap}. The proofs rely on the well-known property of the Lasso optimization problems, namely that if the sign pattern of the solution is known, then we can get the solution in closed form.  
  
  \subsection{Optimality Conditions}
  We let denote $\e = \Y - \X \w \in \rb^n$, $\hQ =  \X^\top \X/n \in \rb^{p \times p} $ and $\hq =  \X^\top \e /n \in \rb^p $. 
    First, we can equivalently rewrite \eq{lasso}  as:
\BEQ
\label{eq:lasso-bis} \min_{w \in \rb^p} \textstyle \frac{1}{2} ( w - \w)^\top \hQ ( w - \w) - \hq^\top ( w - \w) + \mu_n  \| w\|_1.
\EEQ
The optimality conditions for \eq{lasso-bis} can be written in terms of  the sign pattern $s = s(w) = \sign(w)$ and the sparsity pattern $J  = J(w) = \{ j, \ w_j \neq 0 \}$~\cite{yuanlin}:
\BEA
 \nonumber && \!\!\!\! \!\!\!\! \| 
  ( \hQ_{J^c J} \hQ_{J J}^{-1}  \hQ_{J \J}  -  \hQ_{J^c \J} ) \w_\J  +  (\hQ_{J^c J}\hQ_{J J}^{-1} \hq_J-  \hq_{J^c} )   \\
 \label{eq:opt1-bis}
 & & \hspace*{2cm} + \mu_n \hQ_{J^c J} \hQ_{J J}^{-1}  s_J  
   \  \|_\infty \leqslant \mu_n , \\
\label{eq:opt2-bis}  &&   \!\!\!\!    \sign  ( \hQ_{J J}^{-1}  \hQ_{J \J} \w_\J +  \hQ_{J J}^{-1} \hq_J - \mu_n \hQ_{J J}^{-1}  s_J  ) = s_J.
\EEA 
In this paper, we focus on regularization parameters $\mu_n$ of the form $\mu_n = \mu_0 n^{-1/2}$. The main idea behind the results is to consider that $(Q,q)$ are distributed according to their limiting distributions, obtained from the law of large numbers and the central limit theorem, i.e., $Q$ converges to $\Q$ a.s.~and   $n^{1/2} q$ is asymptotically normally distributed with mean zero and covariance matrix $\sigma^2 \Q$. When assuming this, Propositions~\ref{prop:prop1} and~\ref{prop:prop2} are straightforward. The main effort is to make sure that we can safely replace $(Q,q)$ by their limiting distributions.
The following lemmas give sufficient conditions for correct estimation of the signs of variables in $\J$  and for selecting a given pattern $s$ (note that all constants could be expressed in terms of $\Q$ and $\w$, details are omitted here):
\begin{lemma}
\label{lemma:lemma1}
Assume \hypref{inv} and $\| Q - \Q \|_2  \leqslant \lmin(\Q)/2$. Then $\sign(\hat{w}_\J)  \neq 
\sign(\w_\J)$ implies
$
\|\Q^{-1/2} q\|_2 \geqslant C_1 - \mu_n C_2
$,
where
$C_1, C_2 >0$.
\end{lemma}
\begin{lemma}
\label{lemma:lemma2}
Assume \hypref{inv}  and let $s \in \{-1,0,1\}^p$ such that
$s_\J = \sign(\w_\J)$. Let $J = \{ j, s_j \neq 0\} \supset \J$. Assume
\BEQ
\label{eq:l1}
 \| \Q - Q\|_2 \leqslant  \min \left\{ \eta_1 , \lmin(\Q)/2 \right\},
 \EEQ

\BEQ
\label{eq:l2}
  \|\Q^{-1/2} q\|_2 \leqslant \min\{\eta_2,C_1 - \mu_n C_4\}, \EEQ

$$
 \| \Q_{J^c J}\Q_{J J}^{-1} \hq_J-  \hq_{J^c}  - \mu_n \Q_{J^c J}\Q_{J J}^{-1} s_J
\|_\infty  \leqslant \mu_n \hspace*{1cm}  $$

\BEQ
\label{eq:l3}
\hspace*{3.5cm}
 - C_5 \eta_1 \mu_n - C_6 \eta_1 \eta_2,
\EEQ

\BEQ
\label{eq:l4}
\forall i \in J \backslash \J,   s_i \left[ Q_{JJ}^{-1}( q_J \!- \!\mu_n s_J) \right]_i
\! \geqslant \! \mu_n C_7 \eta_1\! + \!C_8 \eta_1 \eta_2,
\EEQ

 with   
$C_4,C_5,C_6,C_7,C_8$ are positive constants.
Then  $\sign(\hat{w}) = \sign(\w)$.
\end{lemma}
Those two lemmas are interesting because they relate optimality of certain sign patterns to quantities from which we can derive concentration inequalities.

\subsection{Concentration Inequalities}
\label{sec:concentration}
Throughout the proofs, we need to provide upper bounds on the following quantities
$\P (\|\Q^{-1/2} q\|_2 > \alpha)$ and $\P (\| Q - \Q \|_2 > \eta)$.
We obtain, following standard arguments~\cite{concentration}:
 if $\alpha  <  C_9 $ and $\eta<C_{10}$ (where $C_9, C_{10}>0$ are constants),
$$ \textstyle
\P (\|\Q^{-1/2} q\|_2 > \alpha) \leqslant 4p \exp \left( - \frac{n\alpha^2}{2pC_9}\right).
 $$
   $$\textstyle
\P (\| Q - \Q \|_2 > \eta) 
\leqslant  4p^2
\exp \left( - \frac{n\eta^2}{2p^2C_10}\right).
$$
We also consider multivariate \emph{Berry-Esseen inequalities} \cite{bentkus}; the probability
$ \P ( n^{1/2} \hq \in \mathcal{C} )
$
can be estimated as
$
  \P (  t  \in \mathcal{C} )
$
where $t$ is normal with mean zero and covariance matrix
$\sigma^2 \Q$. The error is then \emph{uniformly} (for all convex sets $\mathcal{C}$) upperbounded
by:
$$ 400 p^{1/4} n^{-1/2} \lmin(\Q)^{-3/2} \E |\varepsilon|^3 \| X\|_2^3 = C_{11} n^{-1/2}.
$$

\subsection{Proof of Proposition \ref{prop:prop1}}
By Lemma~\ref{lemma:lemma2}, for any given $A$, and $n$ large enough, the probability that the sign is different from $s$ is upperbounded by

$$ \textstyle \P \!\left(\! \|\Q^{-1/2} q\|_2 \!> \!\frac{A (\log n)^{1/2} }{n^{1/2}}
\!\right)
+ \P\!\left(\! \| \Q - Q \|_2 \!>\!  \frac{A(\log n)^{1/2}}{n^{1/2}}\!\right)
$$

$$ + \P \left\{ t \notin \mathcal{C}(s,\mu_0( 1 - \alpha ) ) \right\} + 2 C_{11} n^{-1/2},
$$

where $\mathcal{C}(s,\beta) $ is the set of $t$ such that (a)
$\| \Q_{J^c J}\Q_{J J}^{-1} t_J-  t_{J^c}  - \beta  \Q_{J^c J}\Q_{J J}^{-1} s_J
\|_\infty  \leqslant \beta  
$ and (b) for all $
  i \in J \backslash \J,   s_i \left[ Q_{JJ}^{-1}( t_J - \beta s_J) \right]_i
 \geqslant 0$. Note that here $\alpha = O( (\log n)n^{-1/2})$ tends to zero and that we have:
 $\P \left\{ t \notin \mathcal{C}(s,\mu_0( 1 - \alpha ) ) \right\} 
\leqslant \P \left\{ t \notin \mathcal{C}(s,\mu_0  ) \right\} + O(\alpha)$. All terms (if $A$ is large enough) are thus $O( (\log n)n^{-1/2})$.

 This shows that $  \P( \sign(\hat{w}) = \sign(\w) ) \geqslant \rho(s,\mu_0) + O( (\log n)n^{-1/2})$ where $ \rho(s,\mu_0) =  \P \left\{ t \in \mathcal{C}(s,\mu_0 ) \right\} \in (0,1)$--the probability is strictly between 0 and 1 because the set and its complement have non empty interiors and the normal distribution has a positive definite covariance
 matrix $\sigma^2 \Q$.  The other inequality can be proved similarly.
Note that the constant in $O( (\log n)n^{-1/2})$ depends on $\mu_0$ but by carefully considering this dependence on $\mu_0$, we could make the inequality uniform in $\mu_0$ as long as $\mu_0$ tends to zero or infinity at most at a logarithmic speed (i.e., $\mu_n$ deviates from $n^{-1/2}$ by at most a logarithmic factor). Also, it would be interesting to consider uniform bounds on portions of the regularization path.

\subsection{Proof of Proposition \ref{prop:prop2}}
From Lemma~\ref{lemma:lemma1}, the probability of not selecting any of the variables in $\J$ is upperbounded by
$\P( \|\Q^{-1/2} q\|_2\! > \!C_1 - \mu_n C_2)
+ \P( \| \Q - Q \|_2 \!>\!  \lmin(\Q)/2)
$,
which is straightforwardly upper bounded (using \mysec{concentration})  by a term of the required form.

\subsection{Proof of Proposition \ref{prop:bolasso} }
In order to simplify the proof, we made the simplifying assumption that the random variables $X$ and $\varepsilon$ have compact supports. Extending the proofs to take into account the looser condition that $\|X\|^2$ and $\varepsilon^2$ have non uniformly infinite cumulant generating functions (i.e., assumption \hypref{var}) can be done with minor changes.
The probability that $\bigcap_{k=1}^m J_k$ is different from $\J$ is upper bounded by
the sum of the following probabilities:

  \paragraph{(a) Selecting at least variables in $\J$:}
the probability that for the $k$-th replication, one index in $\J$ is not selected, each of them which is upper bounded by  $\P( \|\Q^{-1/2} q^\ast \|_2 > C_1/2)
+ \P( \| \Q - Q^\ast \|_2 >  \lmin(\Q)/2)$, where $q^\ast$ corresponds to the ghost sample; as common in theoretical analysis of the bootstrap, we relate $q^\ast$ to $q$ as follows:
$ \P( \|\Q^{-1/2} q^\ast \|_2 > C_1/2 )  \leqslant
\P( \|\Q^{-1/2} (q^\ast - q)\|_2 > C_1/4 ) +
\P( \|\Q^{-1/2} q \|_2 > C_1/4)$ (and  similarly for $\P( \| \Q - Q^\ast \|_2 >  \lmin(\Q)/2)$).
Because we have assumed that $X$ and $\varepsilon$ have compact supports, the bootstrapped variables have also compact support and we can use concentration inequalities (given the original variables
$\X$, and also after expectation with respect to $\X$). Thus the probability for one bootstrap replication
 is upperbounded by $B e^{-Cn}$ where $B$ and $C$ are strictly positive constants. Thus the overall contribution of this part is less than $ mB e^{-Cn}$.

\paragraph{(b) Selecting at most variables in $\J$:} the probability that for all replications, the set $\J$ is not exactly selected (note that this is not tight at all since on top of the relevant  variables which are selected with overwhelming probability, different additional variables may be selected for different replications and cancel out when intersecting).
 
Our goal is thus to bound 
$ \E \left\{  \P( \J^\ast \neq \J | \X)^m \right\}$. By previous lemmas, we have that 
$ \P( \J^\ast \neq \J | \X)
$ is upper bounded by
$ \textstyle \P \left( \|\Q^{-1/2} q^\ast\|_2 > \frac{A (\log n)^{1/2} }{n^{1/2}} | \X \right)
+ \P\left( \| \Q - Q^\ast \|_2 >  \frac{A(\log n)^{1/2}}{n^{1/2}} | \X \right)
$
$ \textstyle  + \P ( t^\ast \notin \mathcal{C}(\mu_0 ) | \X ) + 2 C_{11} n^{-1/2}+ O( \frac{
\log n}{n^{1/2}}),
$
where now, given $\X,\Y$, $t^\ast$ is normally distributed with mean $n^{1/2} q$ and 
covariance matrix $\frac{1}{n} \sum_{i=1}^n \varepsilon_i^2 x_i x_i^\top
$. 

The first two terms and the last two ones are uniformly $O( \frac{
\log n}{n^{1/2}})$ (if $A$ is large enough). We then have to consider the remaining term. We have
$\mathcal{C}(\mu_0) = \{ t^\ast \in \rb^p, \|
\Q_{\J^c \J}\Q_{\J \J}^{-1} t^\ast_\J-  t^\ast_{\J^c}  - \mu_0  \Q_{\J^c \J}\Q_{\J \J}^{-1} \s_\J
\|_\infty \leqslant \mu_0 \}$. By Hoeffding's inequality,  we can replace
 the covariance matrix that depends on $\X$ and $\Y$ by $\sigma^2 \Q$, at cost $O(n^{-1/2})$.
We thus have to bound $\P(n^{1/2}q + y \notin \mathcal{C}(\mu_0) | q)$ for $y$ normally distributed and $\mathcal{C}(\mu_0)$ a fixed compact set. Because the set is compact, there exist constants $A,B>0$ such that, if $\|n^{1/2} q\|_2 \leqslant \alpha$ for $\alpha$ large enough, then 
$\P(n^{1/2}q + y \notin \mathcal{C}(\mu_0) | q) \leqslant 1 - A e^{-B\alpha^2}$. Thus, by truncation, we obtain a bound of the form:
$\E \left\{  \P( \J^\ast \neq \J | \X)^m \right\} 
\leqslant (1 - A e^{-B\alpha^2} + F \frac{\log n}{n^{1/2}} )^m + C e^{-B\alpha^2}
\leqslant  \exp(- m A e^{-B\alpha^2} +  m F \frac{\log n}{n^{1/2}} ) + C e^{-B\alpha^2}$, where we have used Hoeffding's inequality to upper bound $\P(
\|n^{1/2} q\|_2 > \alpha)$. By minimizing in closed form with respect to $e^{-B\alpha^2}$, i.e., with
$e^{-B\alpha^2} 
= \frac{F \log n }{A n^{1/2}} + \frac{ \log (mA/C)}{mA}$, 
we obtain the desired inequality.

\bibliography{bolasso}
\bibliographystyle{mlapa}

\end{document}